\documentclass[runningheads]{llncs}

\usepackage{eccv}

\usepackage{eccvabbrv}

\usepackage{graphicx}
\usepackage{algorithm}
\usepackage{algpseudocode}
\usepackage{array} 
\usepackage{amssymb}
\usepackage{booktabs}
\usepackage{array,makecell}
\usepackage{caption}
\usepackage{wrapfig,lipsum}
\usepackage{enumitem}
\usepackage{amsmath}
\usepackage{bm}
\usepackage{arydshln}

\usepackage[accsupp]{axessibility}  % Improves PDF readability for those with disabilities.

\usepackage{hyperref}

\usepackage{orcidlink}

\begin{document}

\title{PILoRA: Prototype Guided Incremental LoRA for Federated Class-Incremental Learning} 

\titlerunning{PILoRA}

\author{Haiyang Guo\inst{1,2} \and
Fei Zhu\inst{3} \and
Wenzhuo Liu\inst{1,2} \and \\
Xu-Yao Zhang\inst{1,2}\thanks{Corresponding author} \and
Cheng-Lin Liu\inst{1,2}}

\authorrunning{H. Guo et al.}

\institute{MAIS, Institute of Automation, Chinese Academy of Sciences \and
School of Artificial Intelligence, University of Chinese Academy of Sciences \and
Centre for Artificial Intelligence and Robotics, Hong Kong Institute of Science and Innovation, Chinese Academy of Sciences \\
\email{\{guohaiyang2023, zhufei2018\}@ia.ac.cn, liuwenzhuo20@mails.ucas.ac.cn, \{xyz, liucl\}@nlpr.ia.ac.cn}}

\maketitle

\begin{abstract}
    Existing federated learning methods have effectively dealt with decentralized learning in scenarios involving data privacy and non-IID data. However, in real-world situations, each client dynamically learns new classes, requiring the global model to classify all seen classes. To effectively mitigate catastrophic forgetting and data heterogeneity under low communication costs, we propose a simple and effective method named PILoRA. On the one hand, we adopt prototype learning to learn better feature representations and leverage the heuristic information between prototypes and class features to design a prototype re-weight module to solve the classifier bias caused by data heterogeneity without retraining the classifier. On the other hand, we view incremental learning as the process of learning distinct task vectors and encoding them within different LoRA parameters. Accordingly, we propose Incremental LoRA to mitigate catastrophic forgetting. Experimental results on standard datasets indicate that our method outperforms the state-of-the-art approaches significantly. More importantly, our method exhibits strong robustness and superiority in different settings and degrees of data heterogeneity. The code is available at \url{https://github.com/Ghy0501/PILoRA}.
\end{abstract}
\section{Introduction}
\label{sec:intro}

% \vspace{-2mm}
Federated learning (FL)~\cite{mcmahan2017communication} is a novel distributed machine learning paradigm that enables multiple data owners to collaboratively train a shared model while ensuring the privacy of their local data. In recent years, with the increasing emphasis on data privacy in society and the refinement of relevant regulations~\cite{voigt2017eu}, FL has experienced rapid development and has been widely applied in various real-world scenarios ~\cite{khan2021federated,niknam2020federated,rieke2020future, MIR-2022-02-053, MIR-2022-08-261}.

Existing FL methods~\cite{mcmahan2017communication,fedpro,karimireddy2020scaffold} typically rely on the closed-world assumption~\cite{closeword, zhu2024open,cheng2023unified}, that is, the number of classes seen by the model remains the same and unchanged during both the training and testing stages. However, the real world is dynamic and constantly changing, the local clients often need to receive new data to update the global model continuously.  Therefore, \emph{Federated Class Incremental Learning} (\emph{FCIL})~\cite{yoon2021federated, dong2022federated} has been proposed to handle FL tasks in dynamic scenarios. Specifically, each local client can only update the model with new class data at each stage and upload the model parameters to the global server for aggregation, while the global model needs to maintain its discriminative ability for all seen classes. Moreover, the data distribution of local data follows the non-independent and identically distributed (non-IID) assumption~\cite{li2022federated}. 

FCIL presents a more realistic setting for real-world applications, but it also introduces greater challenges, as FCIL needs to simultaneously address the issue of catastrophic forgetting~\cite{mccloskey1989catastrophic, kirkpatrick2017overcoming} and data heterogeneity resulting from non-IID. In existing studies, one way is to store a subset of data from old classes and train them together when learning new tasks~\cite{dong2023no, dong2022federated}, but the amount of old data stored is strictly limited due to privacy protection requirements. An alternative way is to utilize generative models to generate pseudo-samples of old data~\cite{qi2022better, zhang2023addressing}, thus retaining the ability to classify the old classes. However, training a good generator will cause greater computational overhead, and the generator also suffers from catastrophic forgetting. Another direction is to use pre-trained models for fine-tuning~\cite{liu2023fedet, bagwe2023fed}, these methods store different stages of knowledge by maintaining a pool of modules and inserting the corresponding modules at inference time based on the similarity between the input data and the modules. However, storing these modules takes up additional memory space. Although they provide different perspectives for solving FCIL, there is still a large unexplored space, especially the performance of the model under different non-IID settings and different degrees of data heterogeneity. So in order to better address this issue, we start by addressing a fundamental question, \emph{what is the panacea for dealing with CIL and FL?}
% \vspace{2mm}

% \centerline{\textit{What is the panacea for dealing with CIL and FL?}}
% \vspace{1mm}
In other words, if we can find common ground in dealing with CIL and FL, it will be of great help in solving the FCIL problem. Specifically, we observe that: 1)~At the level of feature representation, both FL and CIL tasks require models to learn a intra-class compact and inter-class separable feature representation. On the one hand, for the CIL task, this kind of feature representation helps to reduce the overlap between features of the old and new classes in the deep feature space, thus alleviating the forgetting of old knowledge. For this reason, PASS~\cite{zhu2021prototype} introduces self-supervised learning to assist the learning of feature representations. On the other hand, in the FL task, FedProto~\cite{tan2022fedproto} imposes a constraint on the deep features of samples belonging to the same class. This constraint ensures that these features are proximate to the global prototypes of their respective classes. 2)~At the classifier level, classifier drift is the common enemy of both. In CIL, when learning new classes, the decision boundaries learned from old classes can be severely disrupted, resulting in a significant bias in the classification layer~\cite{zhu2021prototype, zhu2021class}. To address this issue, some methods~\cite{rebuffi2017icarl, belouadah2018deesil, belouadah2019il2m} alleviate bias by directly retaining a portion of the data from previous categories and training the model with new data. Other methods~\cite{zhu2021class, zhu2021prototype} address classifier bias by retaining pseudo features of old classes to assist in training the classification layer; In FL, CCVR~\cite{luo2021no} assesses the similarity among different layers within local models under data heterogeneity and reveals that classifiers exhibit the lowest similarity. This suggests that the classifier of each client models is seriously biased to the local data. Similarly to CIL, \cite{luo2021no, shang2022federated} utilize the statistical information of local features to retrain the classifier on the global server to alleviate classifier bias.

% \begin{figure}[h]
%     \centering
%     \includegraphics[clip, width=0.45\textwidth, height=0.35\textwidth]{figure/source/heatmap.png}
%     \caption{The local data of client 1 and client 2 are non-IID. Resnet50 (the first row) focuses more on local patterns, and the patterns learned are significantly different in the case of data heterogeneity, causing the average model to lose some important information (e.g., fish fins). However, ViT (the second row) is less affected by data heterogeneity, and the averaged model basically retains all the information learned by the local model.
%     }
%     \label{fig:heatmap}
%     \vspace{-3mm}
% \end{figure}

\begin{wrapfigure}{r}{0.47\textwidth}
% \vspace{-2mm}
    \centering
    \includegraphics[clip, width=0.47\textwidth, height=0.35\textwidth]{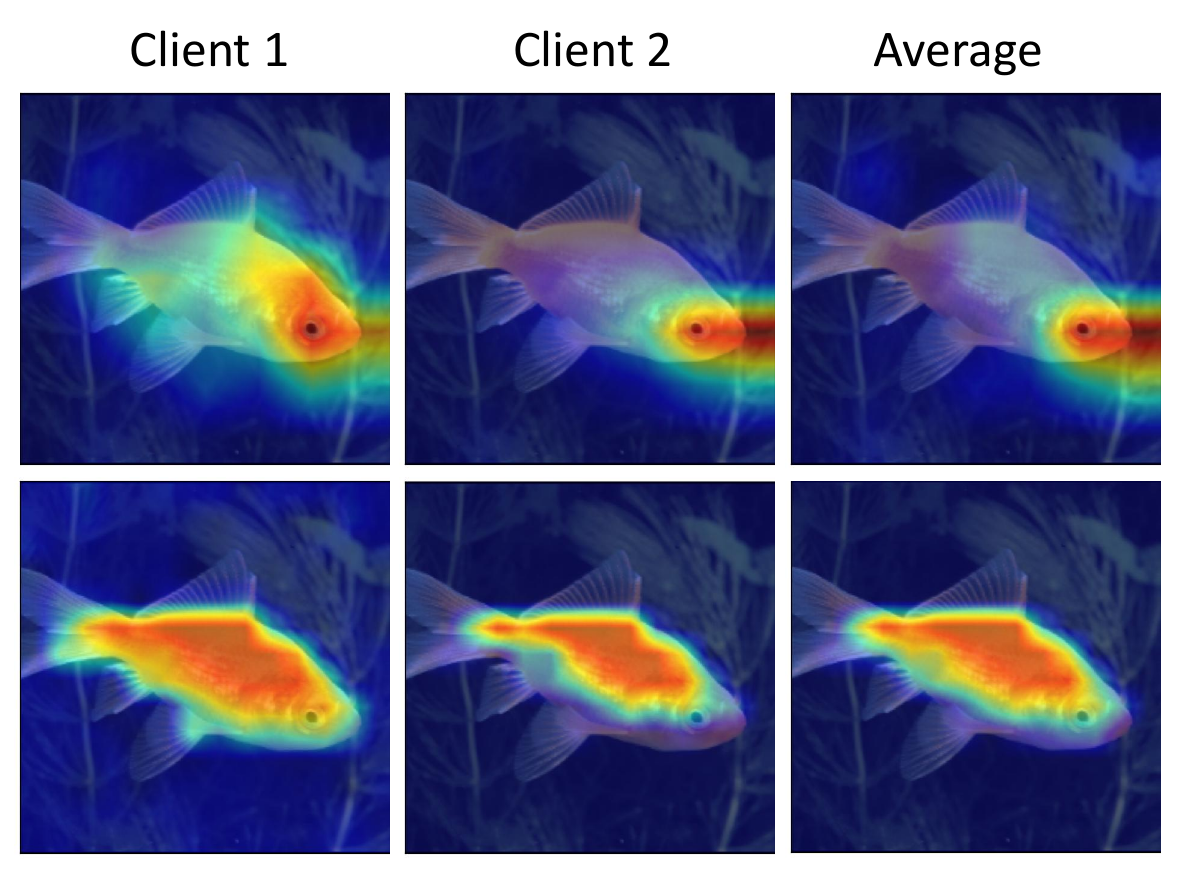}
    \caption{The local data of client 1 and client 2 are non-IID. Resnet50 (the first row) focuses more on local patterns, and the patterns learned are significantly different in the case of data heterogeneity, causing the average model to lose some important information (e.g., fish fins). However, ViT (the second row) is less affected by data heterogeneity, and the averaged model basically retains all the information learned by the local model.
    }
    \label{fig:heatmap}
\end{wrapfigure}
Motivated by the above findings, in this paper, we propose a \textbf{P}rototype Guided \textbf{I}ncremental \textbf{LoRA} (\textbf{PILoRA}) model to tackle the FCIL problem. Specifically, we adopt prototype classifiers in our model to learn intra-class compact and inter-class separable feature representations, which is beneficial for both CIL and FL~\cite{yang2018robust, yang2020convolutional, liu2023class, fedpro}. We use the pre-trained Transformer model~(ViT) as the backbone because the global interactions learned by Transformer are significantly more robust than the local patterns learned by CNN for FL tasks~\cite{qu2022rethinking}~(Fig.~\ref{fig:heatmap}.), and also provide a good representation of the features. Considering the communication cost, we freeze the entire backbone network during training and use LoRA, a Parameter-Efficient Fine-Tuning~(PEFT) method to train the model. To address the catastrophic forgetting problem in FCIL, we propose \emph{incremental LoRA}, which constrains different stages of LoRA to be trained on subspaces orthogonal to each other via orthogonal regularization, and acquires knowledge learned in the past by simple and efficient summation at inference time. To address the classifier bias caused by data heterogeneity, we design a \emph{prototype re-weight} module for the global server, which aggregates local prototypes based on heuristic information between prototypes and corresponding class features. Compared to other existing FCIL methods, our model achieves superior results on standard datasets. Furthermore, we explore the model's performance under different non-IID settings and degrees of data heterogeneity. Experiments demonstrate the robustness of our approach, while other methods suffer from significant declines. 

\noindent
\emph{The main contributions of this paper can be summarized as}:
% \vspace{-4mm}
\begin{itemize}
    \item We propose incremental LoRA, which performs incremental learning on orthogonal subspaces to mitigate catastrophic forgetting and aggregates previous knowledge through simple and efficient parameter summation.
    \item We propose a prototype re-weight module to form global prototypes using heuristic information between each class of local prototypes and the corresponding class of features. Our method effectively addresses classifier bias induced by data heterogeneity without retraining.
    \item We conduct extensive experiments on standard datasets and achieve state-of-the-art performance. Furthermore, in the extreme heterogeneous case, our method still maintains robustness, while all other methods suffer plummet.
\end{itemize}
% \vspace{-6mm}

\section{Related Work}
\label{sec:rl}

\textbf{Federated Class-Incremental Learning:} Research on FCIL has garnered considerable attention in recent years. Dong \etal~\cite{dong2022federated} first introduces the concept of FCIL and proposes several loss functions on the local side and the global server side to alleviate local catastrophic forgetting and global catastrophic forgetting. However, their method uses a rehearsal buffer to store and retain old class data, and additionally design a proxy server to select the best model, resulting in large memory overhead and communication costs. LGA~\cite{dong2023no} extends the work of~\cite{dong2022federated}, but it still belongs to rehearsal-based FCIL. In the more challenging rehearsal-free FCIL problem, generate models are widely adopted to produce synthetic data that aim to mitigate the catastrophic forgetting on local side and global side~\cite{zhang2023target, qi2022better}, but its performance is highly dependent on the quality of synthetic data and will incur additional computational costs. Similar to our work, FedSpace~\cite{shenaj2023asynchronous} designs prototype-based loss to encourage feature vectors of the same class to be close together, while we achieve this goal by using prototype classifier. In contrast to the methods described above, some recent studies~\cite{bagwe2023fed, liu2023fedet} combine pre-trained models with FCIL and achieve higher performance at a smaller communication cost. However, they both adopt a similarity-based selection strategy, which causes additional memory overheads in inference. Besides, they all use supervised pre-training weights, whereas we argue that this may pose privacy concerns because data from downstream tasks may overlap with pre-trained datasets. 
% \vspace{1mm}

\noindent
\textbf{PEFT for Pre-Trained Model:} With the emergence of large-scale pre-trained models~\cite{brown2020language, radford2021learning, kirillov2023segment}, how to effectively fine-tune these models to adapt the downstream tasks has been a focal point of attention. Recently, LoRA~\cite{hu2021lora}, Prompt~\cite{liu2023pre}, and Adapter~\cite{houlsby2019parameter} have emerged as standout techniques and have been widely used in CIL~\cite{wang2022l2p, wang2022dualprompt, smith2023codap, gao2023unified} and FL~\cite{guo2023promptfl, zhao2022reduce, yi2023fedlora} tasks. In FCIL, existing methods~\cite{bagwe2023fed, liu2023fedet} have attempted to combine Prompt and Adapter with pre-trained models. Specifically, they store the knowledge learned at each stage in the parameters of the Prompt or Adapter module and select the appropriate module for the current input to be embedded into the model by a specific similarity computation during the inference time, thus effectively mitigating catastrophic forgetting with a tiny communication costs. However, such similarity matching based approaches undoubtedly introduce inference delays since they require additional similarity computation modules. (In~\cite{liu2023fedet}, they even train a separate CNN to compute the similarity.) Furthermore, they need a additional memory space in the global server to set up a parameter pool to store the module parameters.
% \vspace{-4mm}

\section{Preliminaries}
\label{sec:prelim}
% \vspace{-1mm}

In FCIL setting, each client has a local stream dataset $D_k=\{D_{k}^{t}\}_{t=1}^{T}$, where $D_{k}^{t}=\{\mathbf{X}_{k}^{t}, \mathbf{Y}_{k}^{t}\}=\{x_{k,i}^{t}, y_{k,i}^{t}\}_{i=1}^{N_{t}}$ is the dataset of the $k$-th client on task $t$. Dataset $D_{k}^{t}$ contains $N_{k}^{t}$ training samples and their label $\mathbf{Y}_{k}^{t}\in C_{k}^{t}$, where $C_{k}^{t}$ is the class set of the $k$-th client in task $t$. In particular, the distribution of different clients $k$ under the same task is non-IID and the class sets of different tasks $t$ are disjoint. For local client, the objective is to minimize a pre-defined loss function $\mathcal{L}$ on current dataset $D_{k}^{t}$, while avoiding interference with and possibly enhancing the knowledge acquired from previous learning stages: 
\begin{equation}
    \underset{\boldsymbol{\omega}_{k}^{t}}{\operatorname{argmin}}\:  \mathcal{L}\left(\boldsymbol{\omega_{k}^{t}};\: \boldsymbol{\omega^{t-1}}, \mathbf{X}_{k}^{t}, \mathbf{Y}_{k}^{t}\right),
\end{equation}
where $\omega_{k}^{t}$ is the parameters of $k$-th local model, and $\omega^{t-1}$ is the global model at previous task. Then, the server updates the global model $\omega^{t}$ by aggregating all uploaded parameters as follows:
\begin{equation}
    \boldsymbol{\omega^{t}}=\sum_{k=1}^{K} \gamma_{k} \boldsymbol{\omega}_{k}^{t}, 
    ~\operatorname{where} \gamma_{k}=\frac{N_{k}^{t}}{\sum_{k'} N_{k'}^{t}}.
\end{equation}

The global model aims to correctly classify the test samples of all seen classes and solve the problem of data heterogeneity at low communication cost.
% \vspace{-3mm}

\section{Our method}
\label{sec:method}

% \vspace{-2mm}
\subsection{Incremental LoRA for Pre-Trained Model}
In FCIL, storing the knowledge of different stages in different modules through PEFT can effectively mitigate the catastrophic forgetting~\cite{bagwe2023fed, liu2023fedet}, but additional similarity computation units are required in the inference time in order to select the appropriate module to embed into the model based on the inputs, which leads to inference delay and additional memory overhead. In order to learn an end-to-end global model without taking up extra storage space, we intuitively believe that modules storing knowledge from different stages can be organically combined into one module that embeds knowledge from all stages. Accordingly, we propose \emph{Incremental LoRA} to effectively address catastrophic forgetting in FCIL. On the one hand, LoRA has the natural advantages of low inference latency and more stable training~\cite{hu2021lora}, on the other hand, Inspired by~\cite{wang2023orthogonal, wang2022l2p}, we introduce the orthogonality loss to constrain LoRA to learn new knowledge in a subspace orthogonal to past tasks, thus better preserving the knowledge of the old classes.

\begin{figure*}[t]
    \centering
    \includegraphics[width=0.95\textwidth, height=0.55\textwidth]{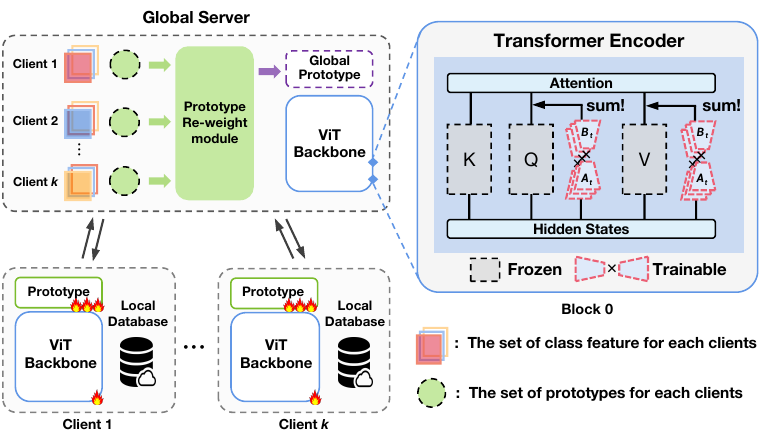}
    \caption{Illustration of PILoRA for FCIL. The client fine-tunes LoRA and prototypes for each class using a local dataset. Upon upload, the global server aggregates the LoRA uploaded by different clients, and applies a prototype re-weight module before re-sending them to each client.
    }
    \label{fig:framework}
\end{figure*}

Specifically, we define the initialization parameters of the pre-training model as $\mathbf{W}\in\mathbb{R}^{d \times k}$, and $\Delta \mathbf{W_t}$ represents the parameter to be updated for task $t$. then, the model's updates in different incremental learning stages can be expressed as:
\begin{equation}
    \mathbf{W}+\Delta \mathbf{W_t}.
\end{equation}

LoRA assumes that the weight changes of large-scale pre-trained models when adapting to downstream tasks occur in a low-rank space:
\begin{equation}
    \mathbf{W}+\Delta \mathbf{W_t}=\mathbf{W}+\mathbf{A}_{t}\mathbf{B}_{t},
\end{equation}
\noindent
where $\mathbf{A}_{t}\in\mathbb{R}^{d \times r}$, $\mathbf{B}_{t}\in\mathbb{R}^{r \times k}$, and $r \ll \min{\{d,k\}}$. At first, $\mathbf{A}_t$ is initialized by a random Gaussian, while $\mathbf{B}_t$ is initialised with zero. Therefore, $\mathbf{B}_t$ can be regarded as the coefficient matrix of $\mathbf{A}_t$~\cite{wang2023orthogonal}. LoRA applies it as a bypass to query and value projection matrices in multi-head attention module, and during adaptation to downstream tasks, only the parameters of $\mathbf{A}_{t}$ and $\mathbf{B}_{t}$ are trainable.

Inspired by~\cite{wang2023orthogonal}, we argue that the parameters of LoRA can be viewed as containers that store different subspaces of task gradients. Thus, learning a series of incremental learning tasks can be viewed as learning a series of LoRA parameters. So how to use these LoRA modules to build a model that can classify all seen classes? An idea is to select appropriate modules to be embedded in the model based on the similarity between different modules and the corresponding inputs~\cite{bagwe2023fed, liu2023fedet}. However, computing similarity requires additional computational resources, which leads to delayed inference. Another idea is that since LoRA is essentially two weight matrices, it is possible to concat all LoRA parameters prior to stage $t$ into a new matrix to gain knowledge of all previous stages~\cite{wang2023orthogonal}. In particular, at stage $t$, the knowledge learned in the previous $t-1$ stages are stored in $\mathbf{A}_{1:t-1}=[\mathbf{A}_1,\dots,\mathbf{A}_{t-1}]$ and $\mathbf{B}_{1:t-1}=[\mathbf{B}_1,\dots,\mathbf{B}_{t-1}]$. To obtain the weight matrix, we can sequentially concat these LoRA parameters into a bigger module:
% \vskip -0.15 in
\begin{equation}
    \mathbf{W}+\Delta \mathbf{W_t}=\mathbf{W}+\mathbf{\widetilde{A}}_{t}\mathbf{\widetilde{B}}_{t},
\end{equation}
% \vskip -0.1 in
\noindent
where $\mathbf{\widetilde{A}}_{t} = \operatorname{concat([\mathbf{A}_1,\dots,\mathbf{A}_{t}])}$, $\mathbf{\widetilde{B}}_{t} = \operatorname{concat([\mathbf{B}_1,\dots,\mathbf{B}_{t}])}$. However, this will cause the parameters of LoRA in the global model to increase with the number of incremental learning tasks. Therefore, in order to ensure consistency between the local model and the global model, we are inspired by Task Arithmetic~\cite{ortiz2024task, ilharco2022editing, chitale2023task} in model edit and propose to integrate the LoRA parameters at different stages by \textbf{summation}:

\begin{equation}
    \mathbf{W}+\Delta \mathbf{W_t}=\mathbf{W}+\mathbf{\overline{A}}_{t}\mathbf{\overline{B}}_{t},
    ~\operatorname{where} \mathbf{\overline{A}}_{t}=\sum_{i=1}^{t} \mathbf{A}_i, ~\mathbf{\overline{B}}_{t}=\sum_{i=1}^{t} \mathbf{B}_i.
\end{equation}

Specifically, distinct directions in the weight space correspond to various localized regions in the input space~\cite{ortiz2024task}. Consequently, a linear combination of these diverse directions encoded within the pre-trained weights enables the model to effectively discriminate between different inputs. In our method, The training data for different incremental tasks are disjoint from each other, so it is reasonable to directly sum the LoRA parameters corresponding to different tasks. Besides, Task Arithmetic also point out that orthogonality between different task parameter vectors helps to better integrate different task. So we propose a orthogonal regularization to achieve this by constraining the parameters of LoRA to be orthogonal to previous tasks:
\begin{equation}
    l_{ort}(\mathbf{A}_i, \mathbf{A}_t)=\sum_{i=1}^{t-1}|\mathbf{A}_{i}^{T}\cdot \mathbf{A}_{t}|.
\end{equation}

Meanwhile, the orthogonal regularization encourages the learning of distinct tasks along orthogonal directions, thereby effectively reducing the spatial overlap between them. This approach will be proved to be advantageous in mitigating catastrophic forgetting, as it helps preserve the knowledge learned from previous tasks while adapting to new ones.

\subsection{Prototype learning with Prototype Re-weight}
\textbf{Prototype Learning:} As mentioned before, a feature representation that is separable intra-class and compact inter-class is helpful for FCIL tasks. Therefore, the model not only needs to correctly classify known classes but also needs to model the distribution of known classes in the feature space. In open set recognition, CPN~\cite{yang2020convolutional} designs a discriminative loss and generative loss for prototype learning to constrain the range of known classes in the feature space, thus reserving space for samples from unknown classes. Inspired by it, we introduce prototype learning in FCIL. In particular, we set a prototype $m=\{m_{i}|i=1,2,\dots,C\}$ for each class, where $m_i\in\mathbb{R}^{d}$, and the dimensions $d$ of each prototype are the same as the dimensions of the final deep feature space.

To shorten the distance between the class feature and the corresponding prototype, we apply the distance-based cross entropy (DCE) discriminative loss. Given a sample $(x,y)$, DCE uses the distance between sample features $f_{\theta}(x)$ and prototype $m_{i}$ to represent the probability of belonging to class $i$. Considering the normalization of probability, DCE adopts the softmax operation:
\begin{equation}
    p(x\in m_{i}|x)=\frac{\exp(-\delta\cdot\|f_{\theta}(x)-m_{i}\|_{2}^{2})}{\sum_{j=1}^{C}\exp(-\delta\cdot\|f_{\theta}(x)-m_{j}\|_{2}^{2})},
\end{equation}
\noindent
where $\|f_{\theta}(x)-m_{i}\|_{2}^{2}$ is the Euclidean distance between the feature $f_{\theta}(x)$ of the input sample and the prototype $m_{i}$, and $\delta$ is the temperature scalar controlling the hardness of the class distribution. Hence, the distance-based-cross-entropy loss can be defined as:
\begin{equation}
    l_{dce}((x,y);\theta,m)=-\log(x\in m_{y}|x).
\end{equation}

By minimizing the DCE loss, the distance between the sample's feature and the correct prototype will be smaller than other incorrect prototypes. However, features learned only under the discriminative loss may not be compact, which may lead to an overlap in feature representations between new and old classes. To solve this problem, we introduce prototype learning~(PL) loss~\cite{yang2020convolutional}:
\begin{equation}
    l_{pl}((x,y);\theta,m)=\|f_{\theta}(x)-m_{y}\|_{2}^{2},
\end{equation}
\noindent
PL loss decreases the distance between sample features and the corresponding correct prototype, making the model learn a more compact intra-class distribution. Essentially, PL loss is a maximum likelihood regularization of features $f_{\theta}(x)$ under the Gaussian mixture density assumption~\cite{liu2004discriminative, liu2004effects}. 
% \vspace{3mm}

\noindent
\textbf{Prototype Re-weight:} Considering a specific class $c$, under the non-IID setting, each client $k$ holds a portion of the training data $N_{c,k}$ and $\sum_{k=1}^{K}N_{c,k}=N_{c}$, where $N_{c}$ is the sum of training samples of class $c$. In prototype learning, the prototype $m_{k,c}$ learned by each client can effectively reflect the distribution of class $c$ within its local data. That is, for clients that do not have training samples of class $c$, the distance between their corresponding prototype and the features of class $c$ will be greater than for clients that do have training samples. If we give equal weight to these uploaded prototypes during parameter aggregation, classifier drift may occur, especially when the data heterogeneity is high, potentially resulting in the global model losing discriminative capability for that class.

Therefore, we use the heuristic information contained in the distance between the prototype and the corresponding class average features to design the prototype re-weight module. Specifically, at stage $t$, each client $k$ uploads prototypes $m_{t,k}$ and the set of average features for the class they have learned $\mu_{t,k}$ (zero for classes without samples). On the global server, we first compute the sum of distances between the prototype $m_{t,k,c}$ and the average feature $\mu_{t,i,c}$ uploaded by all clients:
\begin{equation}\label{eq_7}
    d_{t,k,c}=\sum_{i=1}^{K}\|m_{t,k,c}-\mu_{t,i,c}\|_{2}^{2} ,
\end{equation}
\noindent
the value of $d_{t,k,c}$ approximates the distance from the prototype $m_{t,k,c}$ to the overall features of class $c$. A smaller value indicates that the prototype is closer to the features of that class uploaded by all clients and during aggregation, we aim to assign it a higher weight. So we perform max-min normalization on the set $D_{t,c}=\{p_{t,k,c}|k=1,2,\dots,K\}$, where $p_{t,k,c}=\frac{1}{d_{t,k,c}}$:
\begin{equation}\label{eq_8}
    \alpha_{t,k,c}=\frac{p_{t,k,c}-\min D_{t,c}}{\max D_{t,c}- \min D_{t,c}}.
\end{equation}

To meet the normalization requirements of the weights, we perform softmax processing on them and obtain the weight coefficient:
\begin{equation}\label{eq_9}
    \omega_{t,k,c}=\frac{\exp(\eta\cdot\alpha_{t,k,c})}{\sum_{i=1}^{K} \exp(\eta\cdot\alpha_{t,i,c})},
\end{equation}
% \vskip 0.05 in
\begin{algorithm}[t]
    \renewcommand{\algorithmicrequire}{\textbf{Input:}}
    \renewcommand{\algorithmicensure}{\textbf{Output:}}
    \caption{Prototype Re-weight of Task t.}
    \label{alg:reweight}
    \begin{algorithmic}[1]
        \Require
        the number of classes $C_{t}$, prototypes of each client $m_{t,k}$, the set of mean features $\mu_{t,k}$ of each client and the temperature coefficient $\eta$
        \Ensure
        global prototype $m_{t}$
        \For{$c=1\to~C_t$}
            \If{$\exists~\mu_{t,i,c} \neq \textbf{0}, \text{where}~i=1\to~K$}
                \State $\mu_{t,c}^{*} \gets~\mu_{t,c}~\text{retain all non-zero mean features}$
                \State $d_{t,k,c} \gets~\text{\emph{Distance}~($m_{t,k},~\mu_{t,c}^{*}$) via Eq.~(\ref{eq_7})}$
                \State $p_{t,k,c} \gets~\text{take the reciprocal of}~d_{t,k,c}$
                \State $\alpha_{t,k,c} \gets~\text{\emph{Max-minNorm}~($p_{t,k,c}$) via Eq.~(\ref{eq_8})}$
                \State $\omega_{t,k,c} \gets~\text{\emph{Softmax}~($\alpha_{t,k,c},~\eta$) via Eq.~(\ref{eq_9})}$
            \Else
                \State $\omega_{t,k,c} \gets~\text{Assign with average value}~1/C_t$
            \EndIf
            \State $\omega_{t,c} \gets~\text{List}~(\omega_{t,1,c}, \dots, \omega_{t,K,c})$
            \State $m_{t,c} \gets~\text{\emph{Re-weight}~($\omega_{t,c},~m_{t,k}$) via Eq.~(\ref{eq_10})}$
        \EndFor
        \State $\text{Global prototype}~m_t \gets~\text{Concat all}~m_{t,c}$
    \end{algorithmic}
\end{algorithm}

\noindent
where $\eta$ is the temperature coefficient that controls the softness and hardness of weights. Finally, we re-weight all local prototypes corresponding to class $c$ according to the obtained weights to get the global prototype of class $c$:
\begin{equation}\label{eq_10}
    m_{t,c}=\sum_{i=1}^{K}\omega_{t,i,c}\cdot m_{t,i,c}.
\end{equation}

We argue that prototype re-weight module allows the global prototypes $m_{t,c}$ to differentially consider the data distribution information inherent in each local prototype, effectively mitigating the classifier bias issue. Compared with FedProto~\cite{tan2022fedproto}, our method further compresses the representation region of the same class in the feature space, which contributes to mitigate catastrophic forgetting. More importantly, our method does not require additional retraining of the classification layer, resulting in significant advantages in computational costs. Algorithm~\ref{alg:reweight} presents the pseudo code of prototype re-weight module.
% \vspace{-4mm}

\subsection{Integrated Objective of PILoRA}
% \vspace{-1mm}
The loss function of PILoRA can be defined as:
% \vspace{-1mm}
\begin{equation}
    l_{total}=l_{dce}+\lambda\cdot l_{pl}+\gamma\cdot l_{ort},
\end{equation}
\noindent
where $\lambda$ and $\gamma$ are two hyper-parameters, and the overall framework of PILoRA is shown in Fig.~\ref{fig:framework}. In general, our method is simple and effective, showing strong performance in different non-IID settings and degrees of data heterogeneity.
% \vspace{-4mm}

\section{Experiments}
\label{exp}

\subsection{Experimental Setups}
\textbf{Benchmark:} To evaluate the proposed PILoRA, we perform our experiments on two well-known datasets: CIFAR-100~\cite{krizhevsky2009learning} and TinyImageNet~\cite{le2015tiny}, We also test the performance of the model on large-scale datasets, specifically, we randomly select 200 classes from the ImageNet-1k~\cite{deng2009imagenet} as a new dataset. Following the protocols proposed in~\cite{liu2023fedet}, we split 10 incremental stages and only the data from the current stage is available. Besides, to challenge our method, the local dataset of each client is followed by two kinds of non-IID settings: \emph{quantity-based label imbalance} and \emph{distribution-based label imbalance}~\cite{li2022federated} and we denote the degree of heterogeneity of the two settings by $\alpha$ and $\beta$. Details of both settings can be seen in~\cref{sec:detail_noniid}.
% \vspace{1mm}

\noindent
\textbf{Comparison:} We compare our method with exisiting FCIL methods\footnote{GLFC and LGA only perform experiments on quantity-based label imbalance.}: TARGET~\cite{zhang2023addressing}, GLFC~\cite{dong2022federated}, LGA~\cite{dong2023no}. we also adopt several CIL methods: EWC~\cite{kirkpatrick2017overcoming}, LwF~\cite{li2017learning}, iCaRL~\cite{rebuffi2017icarl}, L2P~\cite{wang2022l2p} and FL method FedNCM~\cite{legate2024guiding} in FCIL setting. In addition, we compare the use of cross-entropy loss for optimization during training and the use of class means as classifiers during inference, which we named FedCLM. We explore their performance in different non-IID settings and degrees of data heterogeneity. For a fair comparison, we tune all methods to the same pre-trained model as ours and fine-tune them using LoRA.
% \vspace{1mm}

\noindent
\textbf{Implementation:} Considering privacy issues, we evaluate the performance of our method on \emph{self-supervised} pre-trained weights (Dino~\cite{caron2021emerging}) for ViT-B/16~\cite{dosovitskiy2020image}, this setting is also widely used in CIL task~\cite{liu2024branch, liu2024towards, zhang2023slca}. Considering the trade-off between performance and number of parameters, we insert the LoRA module into only the first block~(See~\cref{sec:lora_pos}) and set $r=4$. We train our models using Adam~\cite{kingma2014adam} with a batch size of 64 and followed by~\cite{zhang2023slca}, we adopt different learning rates for prototypes layer of $2e^{-3}$ and LoRA parameters of $1e^{-5}$ on CIFAR-100, and $5e^{-3}$/~$5e^{-6}$ on TinyImageNet. Moreover, cosine annealing is also used in training processes. We set $\delta=1, \lambda=0.001$, $\gamma=0.5$ and $\eta=0.2$. We initialize 10 local clients to train and upload the parameters at each communication round. The local training epoch is 5 and the communication round is 30.
% \vspace{-3mm}

\begin{table*}[t]
\setlength\tabcolsep{4pt}
\renewcommand\arraystretch{1.1}
\caption{Results (\%) on CIFAR-100. Results are included for 10 tasks (10 classes per task) and under different degree of two non-IID setting.}
\label{tab:cifar100}
\centering
\scalebox{0.85}{
\begin{tabular}{c||c c||c c||c c||c c||c c||c c} 
\hline
\multicolumn{1}{c||}{Non-IID} & \multicolumn{6}{c||}{Quantity-based label imbalance} & \multicolumn{6}{c}{Distribution-based label imbalance} \\ 
\hline
\multicolumn{1}{c||}{Partition} & \multicolumn{2}{c|}{$\alpha=6$} & \multicolumn{2}{c|}{$\alpha=4$} & \multicolumn{2}{c||}{$\alpha=2$} & \multicolumn{2}{c|}{$\beta=0.5$} & \multicolumn{2}{c|}{$\beta=0.1$} & \multicolumn{2}{c}{$\beta=0.05$} \\ 
\hline
Methods & $A_N$ & \multicolumn{1}{c|}{Avg.} & $A_N$ & \multicolumn{1}{c|}{Avg.} & $A_N$ & Avg. & $A_N$ & \multicolumn{1}{c|}{Avg.} & $A_N$ & \multicolumn{1}{c|}{Avg.} & \multicolumn{1}{c}{$A_N$} & \multicolumn{1}{c}{Avg.} \\ 
\hline
\multicolumn{1}{c||}{Joint}  
& 88.6 & \multicolumn{1}{c|}{-} & 84.3 & \multicolumn{1}{c|}{-} & 79.8 & - & 90.1 & \multicolumn{1}{c|}{-} & 87.8 & \multicolumn{1}{c|}{-} & 85.9 & - \\
\multicolumn{1}{c||}{EWC+FL}  
& 57.9 & \multicolumn{1}{c|}{69.1} & 55.9 & \multicolumn{1}{c|}{66.8} & 42.2 & 52.7 & 65.5 & \multicolumn{1}{c|}{77.8} & 57.8 & \multicolumn{1}{c|}{73.2} & 43.5 & 59.2 \\
\multicolumn{1}{c||}{LwF+FL}  
& 57.4 & \multicolumn{1}{c|}{68.8} & 55.1 & \multicolumn{1}{c|}{66.7} & 40.8 & 52.9 & 64.7 & \multicolumn{1}{c|}{77.5} & 54.6 & \multicolumn{1}{c|}{63.3} & 45.7 & 64.5 \\
\multicolumn{1}{c||}{iCaRL+FL}  
& 35.8 & \multicolumn{1}{c|}{56.5} & 37.1 & \multicolumn{1}{c|}{58.9} & 43.4 & 55.3 & 51.3 & \multicolumn{1}{c|}{67.7} & 50.1 & \multicolumn{1}{c|}{65.9} & 44.6 & 63.0 \\
\multicolumn{1}{c||}{L2P+FL}  
& 63.4 & \multicolumn{1}{c|}{65.1} & 59.0 & \multicolumn{1}{c|}{58.2} &  2.6 & 5.6 & 53.9 & \multicolumn{1}{c|}{51.6} & 62.9 & \multicolumn{1}{c|}{71.4} & 38.7 & 32.2 \\
\multicolumn{1}{c||}{FedCLM}  
& 58.9 & \multicolumn{1}{c|}{69.4} & 57.6 & \multicolumn{1}{c|}{67.6} &  44.3 & 57.9 & 66.5 & \multicolumn{1}{c|}{77.4} & 61.0 & \multicolumn{1}{c|}{71.8} & 48.8 & 63.5 \\
\multicolumn{1}{c||}{FedNCM}  
& 65.6 & \multicolumn{1}{c|}{74.4} & 61.9 & \multicolumn{1}{c|}{71.1} &  49.6 & 59.8 & 66.8 & \multicolumn{1}{c|}{77.9} & 62.1 & \multicolumn{1}{c|}{72.4} & 50.9 & 65.9 \\
\multicolumn{1}{c||}{TARGET}  
& 60.9 & \multicolumn{1}{c|}{71.3} & 58.8 & \multicolumn{1}{c|}{69.5} & 45.2 & 56.5 & 66.1 & \multicolumn{1}{c|}{77.8} & 60.5 & \multicolumn{1}{c|}{71.1} & 51.8 & 65.3 \\
\multicolumn{1}{c||}{GLFC}  
& 58.2 & \multicolumn{1}{c|}{70.4} & 53.7 & \multicolumn{1}{c|}{65.9} &  13.1 & 37.7 & - & \multicolumn{1}{c|}{-} & - & \multicolumn{1}{c|}{-} & - & - \\
\multicolumn{1}{c||}{LGA}  
& 64.5 & \multicolumn{1}{c|}{73.6} & 61.1 & \multicolumn{1}{c|}{70.5} &  21.6 & 40.9 & - & \multicolumn{1}{c|}{-} & - & \multicolumn{1}{c|}{-} & - & - \\
\hline
\multicolumn{1}{c||}{\textbf{Ours}}  
& \underline{69.5} & \multicolumn{1}{c|}{\textbf{78.6}} & \underline{65.1} & \multicolumn{1}{c|}{\underline{74.4}} & \underline{54.9} & \underline{62.6} & \underline{68.5} & \multicolumn{1}{c|}{\underline{78.1}} & \underline{63.4} & \multicolumn{1}{c|}{\underline{73.7}} & \underline{54.8} & \underline{67.1} \\
\multicolumn{1}{c||}{\textbf{Ours+HT}}  
& \textbf{69.6} & \multicolumn{1}{c|}{\underline{78.5}} & \textbf{65.8} & \multicolumn{1}{c|}{\textbf{74.8}} & \textbf{56.5} & \textbf{64.3} & \textbf{70.2} & \multicolumn{1}{c|}{\textbf{78.6}} & \textbf{63.6} & \multicolumn{1}{c|}{\textbf{73.8}} & \textbf{57.9} & \textbf{69.2} \\
\hline
\end{tabular}
}
\end{table*}

\subsection{Comparative Results}
We report $A_N(\uparrow)$ and Avg$(\uparrow)$ to evaluate the performance of the methods, where $A_N$ is the accuracy of all seen classes in the final task and Avg. is calculated as the average accuracy of all tasks. Results are shown in \cref{tab:cifar100} and~\cref{tab:tiny}, we can obverse that our method outperforms other comparison methods and demonstrates strong robustness across different non-IID settings. Among FCIL methods, TARGET achieves relatively better performance under different data heterogeneity, while LGA and GLFC show a steep drop in performance and lack of robustness under extreme data heterogeneity. Notably, FedNCM in the FCIL setting even outperforms TARGET in some cases, which suggests that existing FCIL methods lack the handling of data heterogeneity, whereas our approach dramatically improves the performance of the model under different data heterogeneity through a lightweight prototype re-weight module.

The CIL methods also suffer from data heterogeneity in the FCIL setting, with L2P+FL being the most obvious. We believe that there are two main reasons for this, firstly the performance of L2P relies on supervised pre-trained weights~(e.g., ImageNet-1k~\cite{deng2009imagenet}) and its performance degrades significantly when using self-supervised pre-trained weights~\cite{zhang2023slca}. Secondly, L2P needs to choose the appropriate prompt for embedding in the model based on the similarity calculation. However, in cases of extreme heterogeneity, there exists a significant disparity in what each client learns, even at the same stage. This discrepancy prevents the selection of a uniform prompt for reasoning. Instead, our method bypasses redundant similarity calculations or knowledge distillation~\cite{li2017learning} and effectively mitigates catastrophic forgetting directly through the summation of LoRA parameters. In addition, we combine the HeadTune in FedNCM with our prototype classifier, and the performance of our model is further improved by the better initialization provided by HeadTune. This suggests that that there is still a lot of space for improvement in the existing methods in solving the problem of data heterogeneity in FCIL.

\begin{table*}[t]
\renewcommand\arraystretch{1.1}
\setlength\tabcolsep{4.3pt}
\caption{Results (\%) on TinyImageNet. Results are included for 10 tasks (20 classes per task) and under different degree of two non-IID setting.}
\label{tab:tiny}
\centering
\scalebox{0.85}{
\begin{tabular}{c||c c||c c||c c||c c||c c||c c} 
\hline
\multicolumn{1}{c||}{Non-IID} & \multicolumn{6}{c||}{Quantity-based label imbalance} & \multicolumn{6}{c}{Distribution-based label imbalance} \\ 
\hline
\multicolumn{1}{c||}{Partition} & \multicolumn{2}{c|}{$\alpha=12$} & \multicolumn{2}{c|}{$\alpha=8$} & \multicolumn{2}{c||}{$\alpha=4$} & \multicolumn{2}{c|}{$\beta=0.5$} & \multicolumn{2}{c|}{$\beta=0.1$} & \multicolumn{2}{c}{$\beta=0.05$} \\ 
\hline
Methods & $A_N$ & \multicolumn{1}{c|}{Avg.} & $A_N$ & \multicolumn{1}{c|}{Avg.} & $A_N$ & Avg. & $A_N$ & \multicolumn{1}{c|}{Avg.} & $A_N$ & \multicolumn{1}{c|}{Avg.} & $A_N$ & Avg. \\ 
\hline
\multicolumn{1}{c||}{Joint}  
& 83.6 & \multicolumn{1}{c|}{-} & 82.9 & \multicolumn{1}{c|}{-} & 80.2 & - & 84.3 & \multicolumn{1}{c|}{-} & 83.3 & \multicolumn{1}{c|}{-} & 82.8 & - \\
\multicolumn{1}{c||}{L2P+FL}  
& 61.6 & \multicolumn{1}{c|}{58.0} & 49.4 & \multicolumn{1}{c|}{39.3} &  8.2 & 10.2 & 64.2 & \multicolumn{1}{c|}{66.9} & 56.3 & \multicolumn{1}{c|}{52.5} &  51.9 & 43.2 \\
\multicolumn{1}{c||}{FedCLM}  
& 61.6 & \multicolumn{1}{c|}{72.4} & 51.8 & \multicolumn{1}{c|}{60.3} & 45.8 & 56.9 & 66.5 & \multicolumn{1}{c|}{77.4} & 60.4 & \multicolumn{1}{c|}{71.0} & 46.7 & 57.8 \\
\multicolumn{1}{c||}{FedNCM}  
& 71.6 & \multicolumn{1}{c|}{81.6} & 69.5 & \multicolumn{1}{c|}{79.4} & 57.2 & 64.7 & 73.7 & \multicolumn{1}{c|}{81.6} & 70.8 & \multicolumn{1}{c|}{80.4} & 68.4 & 78.0 \\
\multicolumn{1}{c||}{TARGET}  
& 72.6 & \multicolumn{1}{c|}{81.6} & 70.3 & \multicolumn{1}{c|}{79.6} & 63.8 & 73.5 & 71.6 & \multicolumn{1}{c|}{80.9} & 71.0 & \multicolumn{1}{c|}{80.1} & 69.3 & 79.1 \\
\multicolumn{1}{c||}{GLFC}  
& 69.1 & \multicolumn{1}{c|}{77.9} & 61.3 & \multicolumn{1}{c|}{73.5} & 25.1 & 39.4 & - & \multicolumn{1}{c|}{-} & - & \multicolumn{1}{c|}{-} &  - & - \\
\multicolumn{1}{c||}{LGA}  
& 71.3 & \multicolumn{1}{c|}{79.4} & 65.8 & \multicolumn{1}{c|}{75.3} & 36.7 & 48.8 & - & \multicolumn{1}{c|}{-} & - & \multicolumn{1}{c|}{-} &  - & - \\
\hline
\multicolumn{1}{c||}{\textbf{Ours}}  
& \underline{74.4} & \multicolumn{1}{c|}{\underline{81.0}} & \underline{74.3} & \multicolumn{1}{c|}{\underline{80.9}} & \textbf{70.1} & \textbf{77.8} & \underline{74.5} & \multicolumn{1}{c|}{\underline{80.9}} & \textbf{74.3} & \multicolumn{1}{c|}{\textbf{81.0}} & \textbf{73.6} & \textbf{80.2} \\
\multicolumn{1}{c||}{\textbf{Ours+HT}}  
& \textbf{75.0} & \multicolumn{1}{c|}{\textbf{81.3}} & \textbf{74.9} & \multicolumn{1}{c|}{\textbf{81.2}} & \underline{67.9} & \underline{75.6} & \textbf{74.7} & \multicolumn{1}{c|}{\textbf{81.0}} & \underline{74.3} & \multicolumn{1}{c|}{\underline{80.9}} & \underline{73.4} & \underline{80.1} \\
\hline
\end{tabular}}
\end{table*}

\begin{figure}[tb]
  \centering
  \begin{subfigure}{0.58\linewidth}
    \centering
    \includegraphics[clip, width=0.95\textwidth, height=0.4\textwidth]{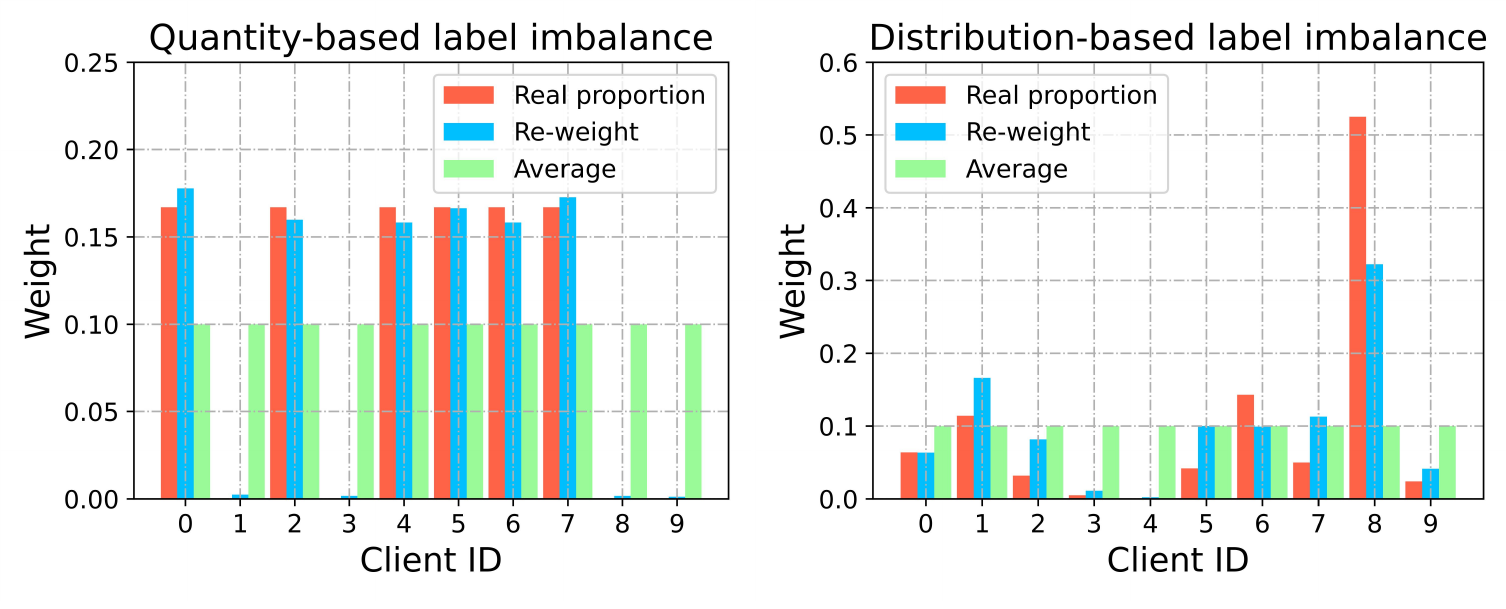}
    \caption{}
    \label{fig:quanzhong}
  \end{subfigure}
  \hfill
  \begin{subfigure}{0.4\linewidth}
  \centering
    \includegraphics[clip, width=0.95\textwidth, height=0.55\textwidth]{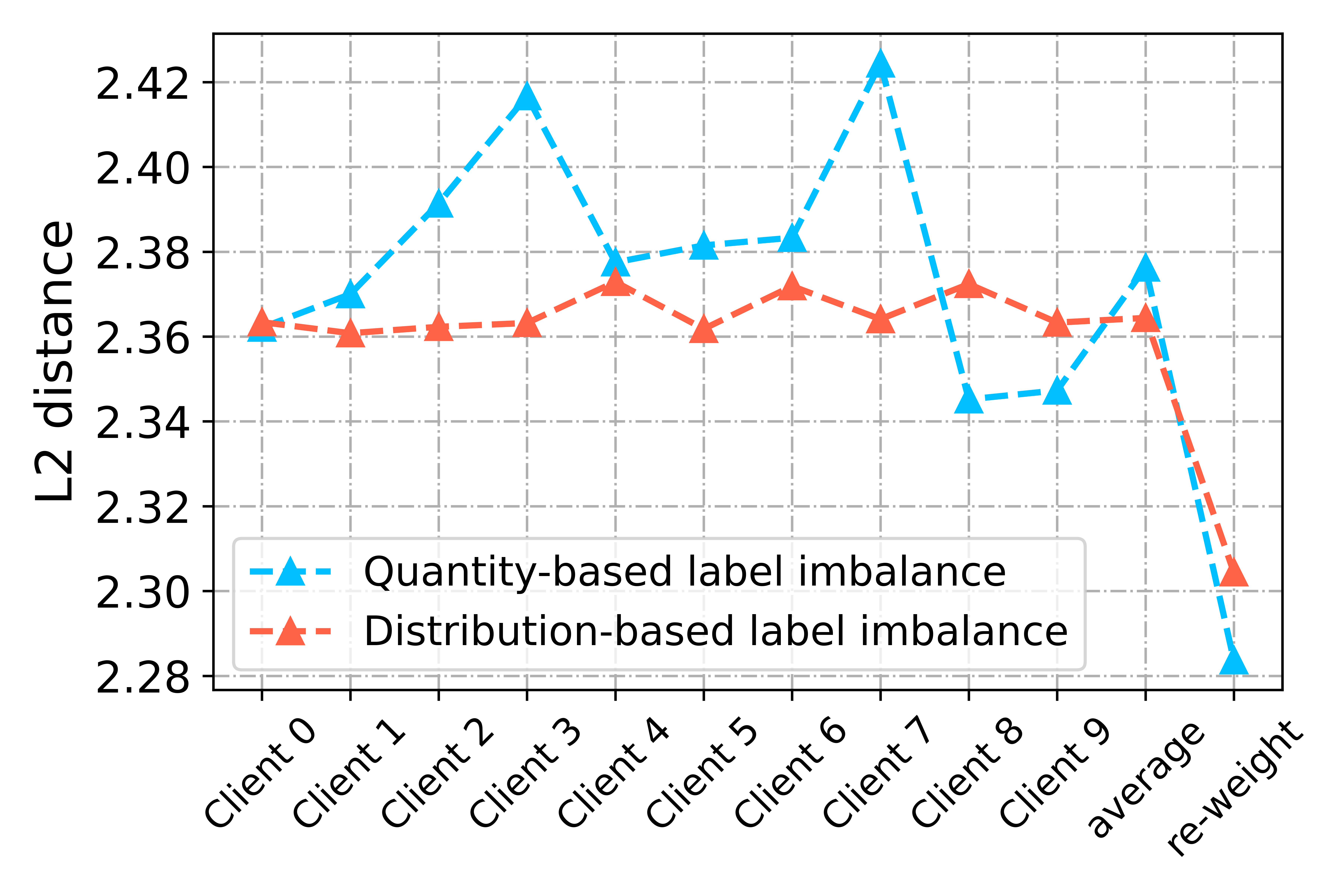}
    \caption{}
    \label{fig:L2-dist}
  \end{subfigure}
  \caption{(a) Comparison of real proportions of the same class among different clients and the weights obtained from Prototype Re-weighting during a local training process. (b) The $L2$ distance between classes feature extracted by different clients and global model with corresponding prototypes.}
  \label{fig:all}
\end{figure}
\noindent
\textbf{Analysis of Prototype Re-weight:} The core idea of our proposed prototype re-weight module is to aggregate prototypes in a manner that closely resembles the true data distribution among clients without disclosing local data information. In \cref{fig:quanzhong}, we show the real proportions of the same class of data across clients and we can observe that the weights calculated through re-weight module align with the distribution of data among different clients. Hence, the global prototypes effectively retain all learned prototype information. In contrast, other methods typically use direct averaging for the classification layer, leading to the fusion of a considerable amount of irrelevant information in the classifier when the data heterogeneity is high, consequently causing classifier bias. In \cref{fig:L2-dist}, we compare the distances between the prototype calculated by both prototype re-weight and average for the global model and the deep features of the test data. It can be seen that our proposed prototype re-weight method effectively ‘takes the best’ of all client uploaded prototypes, so that the global prototype is better adapted to the features of the corresponding class, whereas on average the correct prototype is far away from the features of the corresponding class, especially when the data is very heterogeneous.

\begin{table}[t]
    \centering
        \caption{Ablation Results (\%) on CIFAR-100. Experimental results based on self-supervised pre-trained weights. $F_N~(\downarrow)$ is the average forgetting and the calculation process can be referred to~\cite{zhu2021prototype}.}
        \label{tab:ablation}
        \scalebox{0.95}{
        \begin{tabular}{c||c c c||c c c} 
        \hline
        \multicolumn{1}{c||}{Partition} & \multicolumn{3}{c|}{$\alpha=6$} & \multicolumn{3}{c}{$\beta=0.5$} \\ 
        \hline
        Methods & $A_N$ & Avg. & \multicolumn{1}{c|}{$F_N$} & $A_N$ & \multicolumn{1}{c}{Avg.} & $F_N$ \\ 
        \hline
        \multicolumn{1}{c||}{Ours-w/o PR} 
        & 26.8 & 30.9 & \multicolumn{1}{c|}{\textbf{5.5}} & 41.4 & 48.8 & \multicolumn{1}{c}{9.3} \\
        \multicolumn{1}{c||}{Ours-w/o $l_{ort}$} 
        & 68.7 & 77.9 & \multicolumn{1}{c|}{10.3} & 67.1 & 77.0 & \multicolumn{1}{c}{11.1} \\
        \multicolumn{1}{c||}{Ours-w/o LoRA (fintune)} 
        & 11.3 & 29.5 & \multicolumn{1}{c|}{7.3} & 13.3 & 29.9 & \multicolumn{1}{c}{\textbf{8.9}} \\
        \multicolumn{1}{c||}{Ours-w/o LoRA (frozen)} 
        & 68.6 & 77.7 & \multicolumn{1}{c|}{10.0} & 67.8 & 77.2 & \multicolumn{1}{c}{11.6} \\
        \multicolumn{1}{c||}{\textbf{Ours}}  
        & \textbf{69.5} & \textbf{78.6} & \multicolumn{1}{c|}{9.5} & \textbf{68.5} & \textbf{78.1} & \multicolumn{1}{c}{10.6} \\
        \hline
        \end{tabular}}
\end{table}
\subsection{Ablation Study}
To evaluate the effect of each component in PILoRA, we perform the ablation study and show the results of 10 phases set in CIFAR-100 in \cref{tab:ablation}. We can observe that without prototype re-weight~(PR), The classification accuracy $A_N$ and Avg. of the model decrease significantly, which is due to the fact that the global prototype obtained by simply averaging at this point does not represent the information of each class well, which leads to classifier bias. Instead, our proposed method significantly improves the performance of the model by heuristically re-weight the local prototypes. When there is no orthogonal regularization ~($l_{ort}$), all the metrics of the model show a certain decrease, which indicates that there is a partial overlap between the parameter spaces of different incremental tasks. By imposing orthogonal regularization the parameter spaces corresponding to different input spaces can be made to train the model in the direction of orthogonality to each other, which further improves the model performance. We visualize the cosine similarity between LoRA in~\cref{sec:lora_sim}.

To better demonstrate the contribution of LoRA, we compare the effects of fine-tuning the entire backbone and freezing the entire backbone under self-supervised pre-trained weights when $\alpha=6$ and $\beta=0.5$. As can be seen in \cref{tab:ablation}, when fine-tuning the entire backbone, the performance of the model is severely degraded, we believe that this is due to the fact that the model parameters of ViT are too large to achieve a balance between the ability to remember old classes and the ability to discriminate new classes under the constraint of knowledge distillation. Freezing the entire backbone network can retain discriminative ability for old classes, but at this point, the model has too few trainable parameters, with only prototypes used for classification, thereby limiting the expressive capacity of the model. Therefore, to achieve a trade-off between communication cost and model performance, we fine-tune and achieve the best performance with a tiny number of parameters through Incremental LoRA.

\subsection{Further Analysis}

\begin{wrapfigure}{r}{0.5\textwidth}
    \centering
    \includegraphics[clip, width=0.5\textwidth]{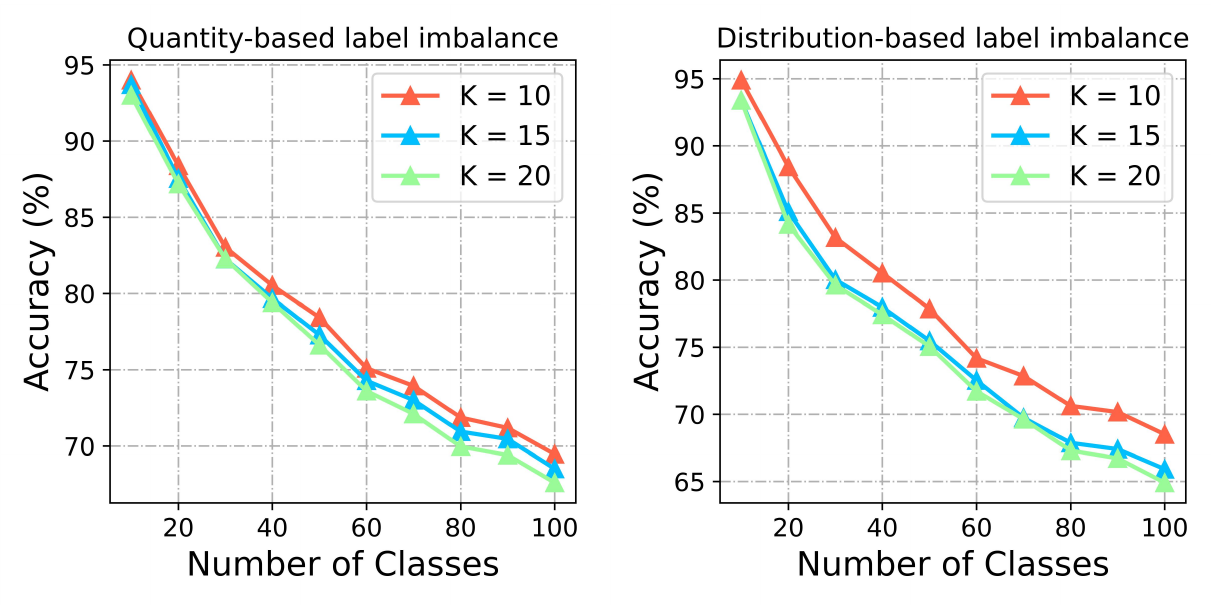}
    \caption{The impact of different number of $K$ on CIFAR-100, where we consider $\alpha=6$ and $\beta=0.5$.}
    \label{fig:k_num}
\end{wrapfigure}

\textbf{Memory usage analysis}. For our PILoRA, in addition to the model for each client, we store the LoRA parameters $A_{q}^{1:t-1}$ and $A_{v}^{1:t-1}$ up to the current stage $t$ to compute the orthogonal regularization. Whereas TARGET requires an additional generator to generate old samples, as well as additional memory space to store the generated images, LGA and GLFC similarly take up additional space to store old class samples. Compared to them, our method takes up very little memory space, storing on average only 0.04\% of the parameters equivalent to the ViT-base.

\noindent
\textbf{Increase the number of local clients~($K$)}. As depicted in \cref{fig:k_num}, we investigat the performance of the model as the number of local clients increase, by respectively setting $K=\{10,15,20\}$. From the results shown in \cref{fig:k_num}, we observe that the model's performance slightly declines as the number of clients increases. We believe that under the same non-IID setting, enlarging the number of clients further exacerbates the heterogeneity among clients, consequently impacting the model's performance. More results can be seen in~\cref{sec:large_scale}.

% \vspace{-3mm}

\section{Conclusion}
\label{con}

In this paper, we propose a simple and effective method of \emph{PILoRA} for FCIL. PILoRA is based on pre-trained ViT models and fine-tunes with a tiny number of parameters using LoRA. To address the catastrophic forgetting problem in FCIL, we propose incremental LoRA, which can efficiently combine different incremental tasks by summing the orthogonal LoRA parameter space; To deal with the classifier bias caused by data heterogeneity, we adopt prototype learning and propose prototype re-weight, which utilize heuristic information between prototypes and features to perform weighted aggregation of global prototypes. Experimental results show that our method achieves state-of-the-art results on standard datasets and maintains robustness under extreme data heterogeneity.

% \newpage
\section*{Acknowledgements}
This work has been supported by the National Science and Technology Major Project (2022ZD0116500), National Natural Science Foundation of China (U20A20223, 62222609, 62076236), CAS Project for Young Scientists in Basic Research (YSBR-083), and Key Research Program of Frontier Sciences of CAS (ZDBS-LY-7004) and the InnoHK program.

\bibliographystyle{splncs04}
\bibliography{main}

\clearpage
\appendix

\section{Details of Non-iid Settings}\label{sec:detail_noniid}
\begin{figure*}[h]
    \centering
    \includegraphics[width=1.0\textwidth, height=0.38\textwidth]{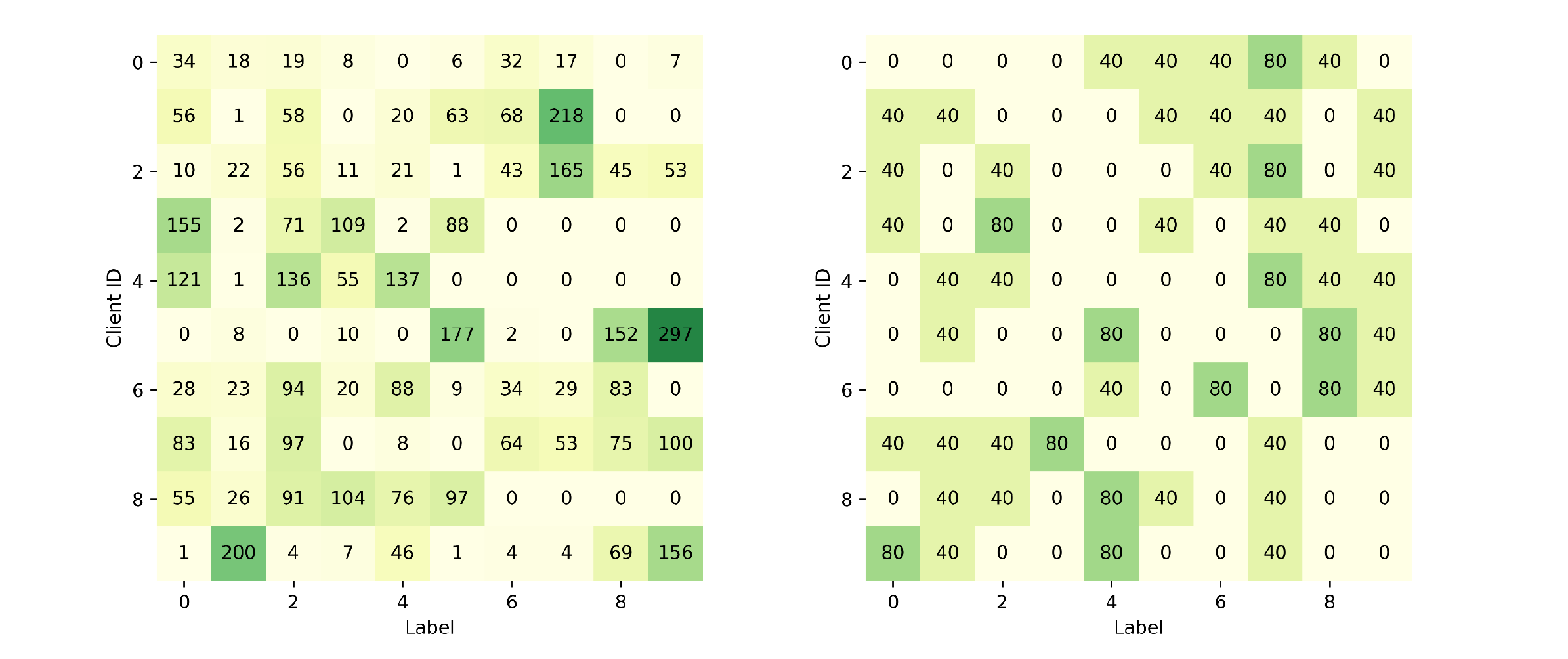}
    \caption{An example of distribution-based label imbalance partition and quantity-based label imbalance partition on CIFAR-100~(10 classes) with $\beta = 0.5$~(left) and $\alpha=6$~(right).
    }
    \label{fig:non_iid}
\end{figure*}

In quantity-based label imbalance, we randomly assign $\alpha$ different label IDs to each client at each stage. Then, for each labeled sample, we randomly and equally distribute it among the clients associated with that label. In distribution-based label imbalance, each client receives a portion of the samples for each label based on the Dirichlet distribution. Formally, we sample $P_k$ from $Dir_N(\beta)$ and allocate a proportion $P_{k,j}$ of class $k$ instances to the client $j$. \cref{fig:non_iid} shows these two partitioning strategies.

\section{Additional Experiments.}

\subsection{Impact of LoRA embedded in different blocks}\label{sec:lora_pos}

\begin{figure*}[h]
    \centering
    \includegraphics[width=1.0\textwidth, height=0.25\textwidth]{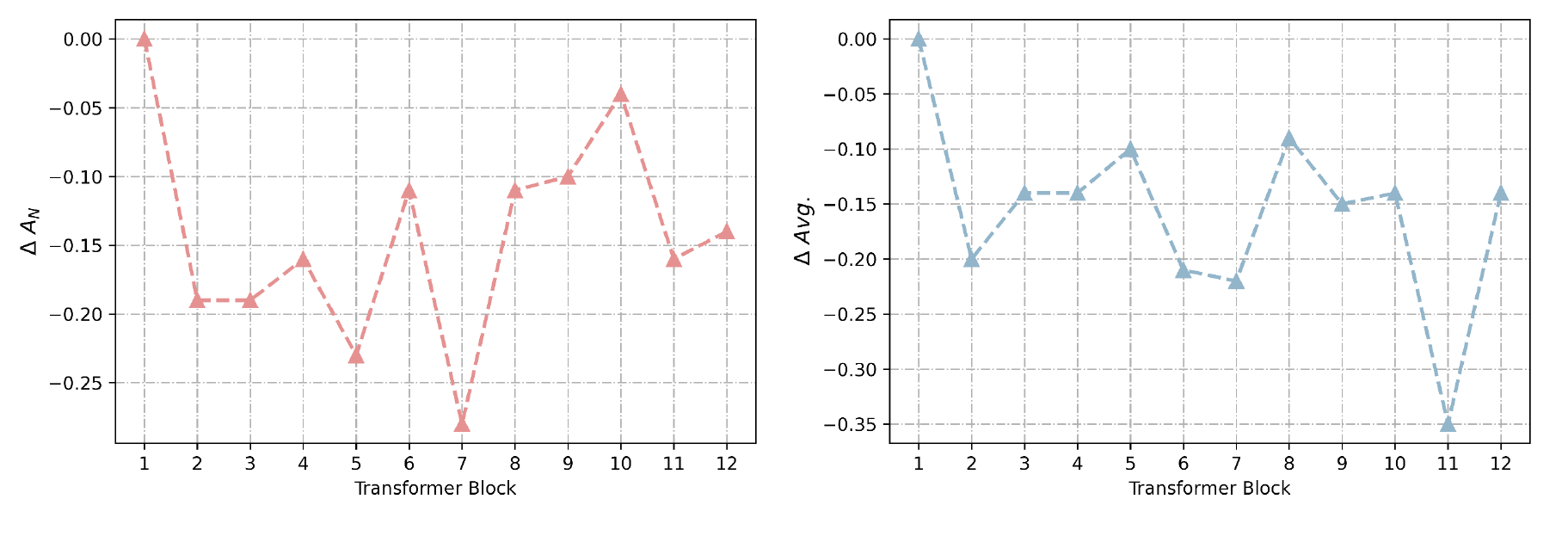}
    \caption{Results of LoRA embedded in different blocks. We visualize the relative accuracy using embedding into the first block as a baseline.
    }
    \label{fig:block}
\end{figure*}

In our method, we embed the LoRA module in the first block of the model. Here we test the results of the LoRA module embedding in each ViT block and compute the relative accuracy~(e.g., $\Delta A_{N}^{i} = A_{N}^{i} - A_{N}^{0}|_{i=1,...,12}$) of embedding in each block versus embedding in the first block. As can be seen in~\cref{fig:block}, embedding LoRA in the first layer consistently outperforms embedding it in any other layers. Therefore, in our method, we fix LoRA to be embedded specifically in the first block.

\subsection{Similarity analysis of LoRA parameters}\label{sec:lora_sim}

\begin{figure*}[h]
    \centering
    \includegraphics[width=1.0\textwidth, height=0.39\textwidth]{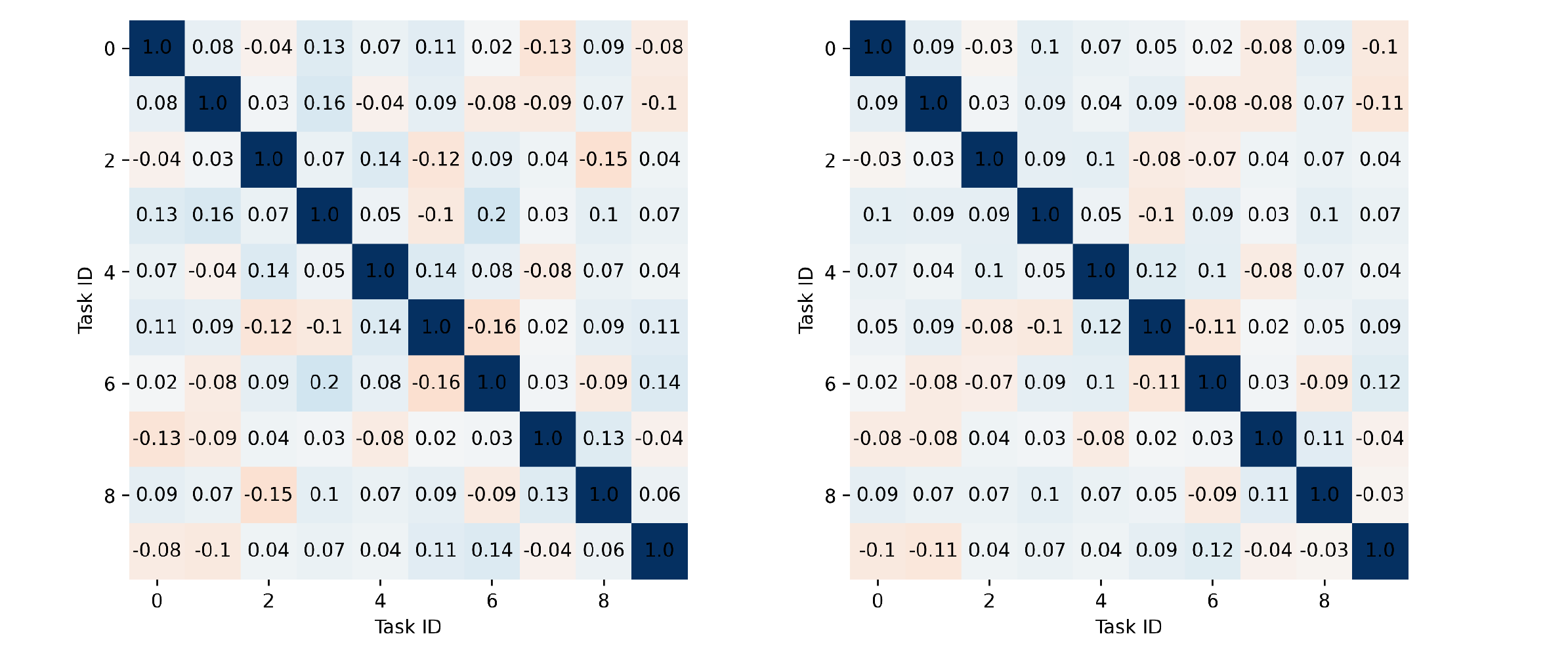}
    \caption{LoRA cosine similarity visualization. Left: without $l_{ort}$; Right: with $L_{ort}$.
    }
    \label{fig:ort}
\end{figure*}

In order to better demonstrate the role of orthogonal regularization, we compute the average cosine similarity of LoRA parameters between different stages, and the results are shown in~\cref{fig:ort}. It can be seen that under the effect of orthogonal regularization, the cosine similarity between LoRAs at different stages is relatively low, indicating that the parameter space is closer to orthogonality, and therefore mitigates catastrophic forgetting.

\subsection{Large-scale experiments}\label{sec:large_scale}

\begin{table*}[h]
\caption{Experiments on large scale dataset~(Imagenet-200).}
\renewcommand\arraystretch{1.0}
\centering
\label{tab:imagenet1k}
\scalebox{1.0}{
\begin{tabular}{c||c c c||c c c} 
        \hline
        \multicolumn{1}{c||}{} & \multicolumn{3}{c||}{\textbf{10 Tasks}} & \multicolumn{3}{c}{\textbf{20 Tasks}} \\
        \hline
        Methods & $A_N$ & Avg. & $F_N$ & $A_N$ & Avg. & $F_N$ \\
        \hline
        L2P+FL & 33.5 & 53.2 & 12.5 & 15.6 & 39.8 & 16.2 \\
        Ours & \textbf{80.1} & \textbf{83.8} & \textbf{4.0} & \textbf{79.5} & \textbf{83.4} & \textbf{4.6}  \\
        \hline
\end{tabular}}
\end{table*}

We also test the performance of the model on large-scale datasets, specifically, we randomly select 200 classes from Imagenet-1k as a new dataset and use self-supervised pre-trained weights. As can be seen in \cref{tab:imagenet1k}, in quantity-based label imbalance, our model still maintains good performance on large-scale datasets. In addition, we also test the performance of the model on longer incremental phases~(20 tasks), and our method effectively mitigates catastrophic forgetting in the long-phase incremental task compared to L2P+FL.
\end{document}